\documentclass[11pt,a4paper]{article}
\usepackage[hyperref]{main}
\usepackage{times}
\usepackage{latexsym}
\usepackage{graphicx}
\usepackage{makecell}
\usepackage{multirow}
\usepackage{tikz}

\usepackage{url}

\aclfinalcopy 

\title{Task-Oriented Language Grounding for Language Input with Multiple Sub-Goals of Non-Linear Order}

\author{Vladislav Kurenkov \\
  Innopolis University \\
  \And
  Bulat Maksudov \\
  Innopolis University \\
  \texttt{\{v.kurenkov, b.maksudov, a.khan\}@innopolis.ru} \\
  \And
  Adil Khan \\
  Innopolis University
  }

\date{}

\begin{document}
\maketitle
\begin{abstract}
In this work, we analyze the performance of general deep reinforcement learning algorithms for a task-oriented language grounding problem, where language input contains multiple sub-goals and their order of execution is non-linear.

We generate a simple instructional language for the GridWorld environment, that is built around three language elements (order connectors) defining the order of execution: one linear -- ``comma`` and two non-linear -- ``but first``, ``but before``. We apply one of the deep reinforcement learning baselines -- Double DQN with frame stacking and ablate several extensions such as Prioritized Experience Replay and Gated-Attention architecture.

Our results show that the introduction of non-linear order connectors improves the success rate on instructions with a higher number of sub-goals in 2-3 times, but it still does not exceed 20\%. Also, we observe that the usage of Gated-Attention provides no competitive advantage against concatenation in this setting.
\end{abstract}

\section{Introduction}
Task-oriented language grounding refers to the process of extracting semantically meaningful representations of language by mapping it to visual elements and actions in the environment in order to perform the task specified by the instruction \cite{chaplot_gated-attention_2017}.

Recent works in this paradigm focus on wide spectrum of natural language instructions' semantics: different characteristics of referents (colors, relative positions, relative sizes), multiple tasks, and multiple sub-goals \cite{chaplot_gated-attention_2017, hermann_grounded_2017, yu_deep_2017,yu_interactive_2018,yu_guided_2018, chevalier-boisvert_babyai:_2018, oh_zero-shot_2017}.

In this work, we are interested in the language input with the semantics of multiple sub-goals, focusing on the order of execution, as the natural language contains elements that may lead to the non-linear order of execution (e.g. ``Take the garbage out, but first wash the dishes``). We refer to this kind of elements as non-linear order connectors in the setting of task-oriented language grounding.

In particular, we want to answer: what is the performance of general deep reinforcement learning algorithms with this kind of language input? Can it successfully learn all of the training instructions? Can it generalize to an unseen number of sub-goals?

To answer these questions, we generate an instruction's language for a modified GridWorld environment, where the agent needs to visit items specified in a given instruction. The language is built around three order connectors: one linear -- ``comma``, and two non-linear -- ``but first``, and ``but before``, producing instructions like ``Go to the red, go to the blue, but first go to the green``. In contrast to \citet{oh_zero-shot_2017}, where the sub-goals are separated in advance and the order is known, in this work, we specifically aim to study whether the agent can learn to determine the order of execution based on the connectors present in the language input.

We apply one of the offline deep reinforcement learning baselines -- Dueling DQN and examine the impact of several extensions such as Gated-Attention architecture \cite{chaplot_gated-attention_2017} and Prioritized Experience Replay \cite{Schaul2016PrioritizedER}.

First, we discover that training for both non-linear and linear order connectors at the same time improves the performance on the latter when compared to training using only linear ones. Second, we observe that the generalization to an unseen number of sub-goals using general methods of deep reinforcement learning is possible but still very limited for both linear and non-linear connectors. Third, we find that there is no advantage of using gated-attention against simple concatenation in this setting. And fourth, we observe that the usage of prioritized experience replay in this setup may be enough to achieve the training performance near to perfect. \footnote{Source code and experiments' results available at \url{https://github.com/vkurenkov/language-grounding-multigoal}}


\section{Problem Formulation}
\subsection{Environment}
\begin{figure}
    \centering
    \begin{tikzpicture}[scale=0.65]
        \draw[step=1cm,gray,very thin] (0,0) grid (10,10);
        
        \node at (1.5, 5.5) (agent) {A};
        \draw (agent) circle (10pt);
        
        \node at (6.5, 5.5) (red) {R};
        \draw (6.2, 5.2) rectangle ++(20pt,20pt);
        
        \node at (4.5, 9.5) (blue) {B};
        \draw (4.2, 9.2) rectangle ++(20pt,20pt);
        
        \node at (3.5, 0.5) (green) {G};
        \draw (3.2, 0.2) rectangle ++(20pt,20pt);
    \end{tikzpicture}
    \caption{The modified GridWorld environment. The goal of the agent (A) is to visit items in the right order as specified in a given language input. R -- Red, B -- Blue, G -- Green.}
    \label{fig:environment}
\end{figure}
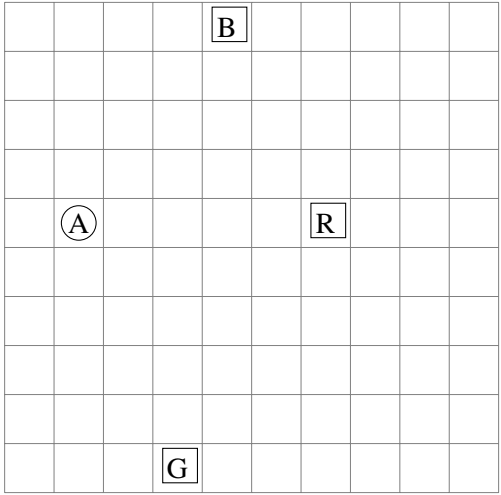

The target environment is the modified version of GridWorld. Figure \ref{fig:environment} illustrates one of the possible layouts. The environment is episodic and bounded by a maximum number of steps which can be restarted indefinitely. The goal of the agent is to visit every object in the right order. For example, in Figure \ref{fig:environment}, if the provided linguistic instruction is ``Go to the blue object, but first go to the green object`` then the agent must visit the green object, and then move to the blue object. If the agent violates the order of visitation or arrives at the wrong object, the episode terminates. 

\subsection{Language}

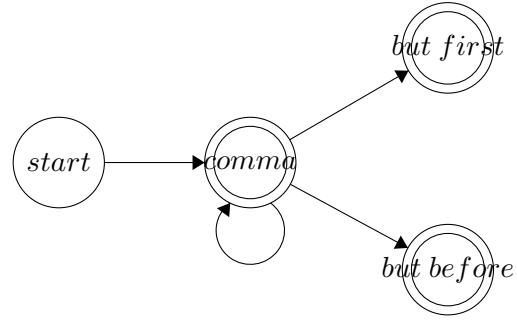
\begin{figure}
    \centering
    \begin{tikzpicture}[scale=0.20]
        \tikzstyle{every node}+=[inner sep=0pt]
        \draw [black] (10.7,-21) circle (3);
        \draw (10.7,-21) node {$start$};
        \draw [black] (23.3,-21) circle (3);
        \draw (23.3,-21) node {$comma$};
        \draw [black] (23.3,-21) circle (2.4);
        \draw [black] (36.3,-13.4) circle (3);
        \draw (36.3,-13.4) node {$but\mbox{ }first$};
        \draw [black] (36.3,-13.4) circle (2.4);
        \draw [black] (36.3,-28.1) circle (3);
        \draw (36.3,-28.1) node {$but\mbox{ }before$};
        \draw [black] (36.3,-28.1) circle (2.4);
        \draw [black] (25.89,-19.49) -- (33.71,-14.91);
        \fill [black] (33.71,-14.91) -- (32.77,-14.89) -- (33.27,-15.75);
        \draw [black] (24.623,-23.68) arc (54:-234:2.25);
        \fill [black] (21.98,-23.68) -- (21.1,-24.03) -- (21.91,-24.62);
        \draw [black] (25.93,-22.44) -- (33.67,-26.66);
        \fill [black] (33.67,-26.66) -- (33.2,-25.84) -- (32.73,-26.72);
        \draw [black] (13.7,-21) -- (20.3,-21);
        \fill [black] (20.3,-21) -- (19.5,-20.5) -- (19.5,-21.5);
    \end{tikzpicture}
    \caption{The finite state automaton describing relations between the order connectors in the target language.}
    \label{fig:fsm_order_connectors}
\end{figure}

The semantics of interest are the presence of multiple sub-goals and the order of their execution. To capture them, we generate an instruction's language, where every instruction describes what objects (referents) the agent should visit and in what order.

The generated language operates on only one task -- ``Go to the ...``, and three referents: red, blue, and green. An instruction may contain multiple sub-goals, and these sub-goals are connected in special ways. In the setting of task-oriented language grounding, we denote the connections between sub-goals that define the order of execution as the order connectors.

We distinguish between linear and non-linear order connectors. The former refers to the connectors that preserve the order as they appear in the language input, e.g. ``Go to the red, go to the blue``, ``comma`` is a linear connector, as the order of execution is the same as the sub-goals ordered in the input. The non-linear connectors may change the order of execution in any way, e.g. ``Go to the red, go to the green, but first go to the blue``, ``but first`` is a non-linear connector as the last sub-goal in the language input should be executed the first.

The generated language contains three order connectors: one linear -- ``comma``, and two non-linear -- ``but first``, ``but before``. The connector ``but before`` swaps the order of two consecutive sub-goals, e.g. ``Go to the red, go to the green, but before go to the blue`` resolves to [red, blue, green]. Figure \ref{fig:fsm_order_connectors} depicts how the language is generated on the level of order connectors. If ``but before`` or ``but first`` connectors are present, we restrict them to be the last order connector in the instruction to avoid ambiguity in the language. The generated language excludes all the instructions that resolve to the visitation of the same item two or more times in a row (e.g ``Go to the red, go to the blue, but first go to the red`` is not allowed).

\subsection{Evaluation}
\begin{table*}
    \centering
    \begin{tabular}{|c|c|c|c|c|}
    \hline
    \textbf{Language Subset} & \textbf{Example Instruction} & \textbf{Example Order} \\\hline
    Comma & Go to the red, go to the blue, go to the green & red, blue, green \\\hline
    Comma-ButFirst & Go to the red, go to the blue, but first go to the green & green, red, blue \\\hline
    Comma-ButBefore & Go to the red, go to the blue, but before go to the green & red, green, blue \\\hline
    \end{tabular} 
    \caption{Example instructions of the target language subsets.}\label{tab:instructions}
\end{table*}
\begin{table*}
    \centering
    \begin{tabular}{|c|c|c|c|}
    \hline
    \textbf{Language Subset} & \textbf{Order Connectors} & \textbf{\# Training} & \textbf{\# Testing} \\\hline
    Comma & comma & 21 & 168\\\hline
    Comma-ButFirst & comma, but first & 39 & 336\\\hline
    Comma-ButBefore & comma, but before & 39 & 336\\\hline
    \end{tabular} 
    \caption{Description of the target language subsets. Comma is a subset of both Comma-ButFirst and Comma-ButBefore.}\label{tab:conjunctions}
\end{table*}
The evaluation is built around instructions' allocation in order to assess the effect of different order connectors. We extract three subsets of the language: Comma, Comma-ButFirst, Comma-ButBefore. The first subset is designated to assess the performance on the language input with linear order connectors only. The goal of the two last subsets is to measure the performance in the presence of non-linear connectors of different type: absolute position change (Comma-ButFirst) and the relative position change (Comma-ButBefore).

\begin{figure}
    \centering
    \includegraphics[width=1.0\columnwidth]{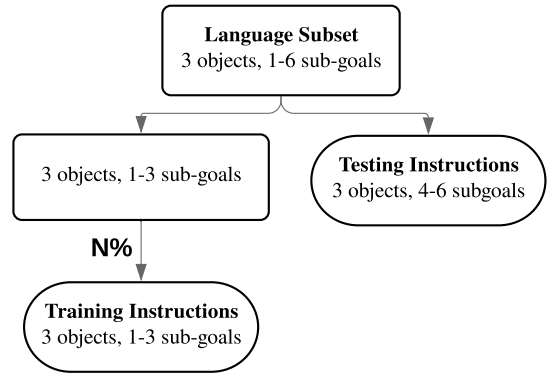}
    \caption{The split of instructions within the generated language subsets. In the end, we obtain two splits: training and testing. The N\% of the instructions with 1-3 subgoals are sampled for training.}
    \label{fig:instr_allocation}
\end{figure}

For every subset, we make two splits, one -- for the evaluation of training performance, the other -- for the evaluation of generalization to a higher number of sub-goals. The splits are obtained as depicted in Figure \ref{fig:instr_allocation}. We limit every subset to include instructions with six sub-goals at max. Then we split it into the training and testing parts. The training part contains all the instructions bounded by 3 sub-goals, and the testing part contains the rest.

Table \ref{tab:conjunctions} describes what order connectors are present in every subset, and how many instructions are in the training and testing splits. The Comma is a subset of both Comma-ButFirst and Comma-ButBefore.

To measure the training performance, we vary the proportion of training instructions from 0.1 to 0.9 with a step of 0.1. The performance on the training instructions is quantified as the success rate on these instructions. To measure the testing performance, we train the models on the biggest proportion of the training instructions.

As for the environment's layout, we randomly sampled only one instance, and utilize it in both training and testing scenarios for all of the algorithms to reduce the computing time, as the layouts generalization is not in the scope of this study.
\section{Methods}
\subsection{Algorithms}
We focus on one of the offline reinforcement learning baseline algorithms - single-actor Dueling DQN (DDQN) \cite{Wang2016DuelingNA}. In particular, we vary two components: the network's architecture and the form of the experience replay.

For the latter, we examine the Prioritized Experience Replay \cite{Schaul2016PrioritizedER}, which is hypothesized to provide a form of implicit curriculum.

For the former, we experiment with Gated-Attention (GA) and Concatenation (Cat) architectures \cite{chaplot_gated-attention_2017}. As the GA was shown to be advantageous for the language input in VizDOOM environment \cite{kempka_vizdoom:_2016, chaplot_gated-attention_2017}. But, the language input in \citet{chaplot_gated-attention_2017} primarily concentrated on different object attributes, we are interested in whether this mechanism will have a positive impact on instructions with multiple sub-goals as well.

The network architecture is as in \citet{chaplot_gated-attention_2017}, but instead of recurrence part, we utilize stacking of 4 previous observations as in \citet{misra_mapping_2017}.

The training loop is organized such that the number of the target network's updates is kept constant for all instructions. Meaning that, instead of random instruction sampling, we iterate over all training instructions in a similar way to \citet{misra_mapping_2017} and only then update our target network.

\subsection{Reward Shaping}
The problem is tackled under the most informative reward shaping scheme. It incorporates the information on how many steps are left to the successful execution of the current sub-goal. 

In order to preserve the optimal policy for the original Markov Decision Process, we apply a potential-based reward transformation \cite{Ng:1999:PIU:645528.657613}.

\section{Results and Discussion}

\begin{table*}
\centering
\begin{tabular}{|c|c|c|c|c|}
\hline
\textbf{Language Subset} & \textbf{DDQN + Cat} & \textbf{DDQN + GA} & \textbf{DDQN + Cat + PER} & \textbf{DDQN + GA + PER} \\
\hline
Comma & 81.1\% & 73.3\% & 89.5\% & \textbf{95.5\%} \\
\hline
\multirow{2}{*}{Comma-ButFirst} & 82.0\% & 75.1\% & \textbf{100.0\%} & 96.2\% \\
\cline{2-5}
& (84.5\%) & (78.1\%) & (100.0\%) & (96.6\%) \\
\hline
\multirow{2}{*}{Comma-ButBefore} & 89.9\% & 72.9\% & \textbf{100.0\%} & 94.4\% \\
\cline{2-5}
& (88.7\%) & (76.3\%) & (100.0\%) & (95.0\%) \\
\hline
\end{tabular} 
\caption{Success rates at training instructions. The scores are averaged over different training proportions. The upper rows show the performance for all training instructions at the corresponding language subset. The bottom rows of Comma-ButFirst and Comma-ButBefore show the performance on the Comma instructions while being trained at Comma-ButFirst or Comma-ButBefore. }\label{tab:training_scores}
\end{table*}

\begin{table*}
\centering
\begin{tabular}{|c|c|c|c|c|}
\hline
\textbf{Language Subset} & \textbf{DDQN + Cat} & \textbf{DDQN + GA} & \textbf{DDQN + Cat + PER} & \textbf{DDQN + GA + PER} \\\hline
Comma & \textbf{10.7\%} & 3.6\% & 6.5\% & 4.8\% \\\hline
\multirow{2}{*}{Comma-ButFirst} & 8.9\% & 7.4\% & \textbf{17.6\%} & 13.4\% \\
\cline{2-5}
& (8.3\%) & (6.5\%) & (16.7\%) & (13.1\%) \\
\hline
\multirow{2}{*}{Comma-ButBefore} & 13.4\% & 7.7\% & \textbf{17.6\%} & 12.5\% \\
\cline{2-5}
& (13.7\%) & (8.3\%) & (20.8\%) & (12.5\%) \\
\hline
\end{tabular} 
\caption{Success rates at testing instructions (higher number of sub-goals). The scores are taken for the runs with the biggest training proportions. The bottom rows of Comma-ButFirst and Comma-ButBefore show the performance on the Comma instructions while being trained at Comma-ButFirst or Comma-ButBefore.}\label{tab:testing_scores}
\end{table*}

The training and testing performance of the Dueling DQN algorithm with different extensions can be found in the Tables \ref{tab:training_scores} and \ref{tab:testing_scores}.

The training at language subsets Comma-ButFirst and Comma-ButBefore substantially improves the training and generalization performance at Comma subset when compared to training only at Comma subset. This is quite an unexpected outcome that suggests that the exposition to the non-linear order connectors may improve the performance for linear connectors.

We notice that the concatenation consistently outperforms the gated-attention mechanism for both training and testing instructions. We suspect that the gated-attention is useful in the scenarios where objects are described in terms of multiple attributes, but it has no to harming effect when it comes to the order connectors.

The frame stacking was enough to achieve the success at training and at some part of the testing instructions. The reason is not clear, but we hypothesize that it can be explained by the lack of the layouts variability and by the offloading mechanism \cite{stefano}. This requires further investigation.

The usage of prioritized experience replay outperforms the simple replay buffer by a big margin -- from 10\% to 20\% success rate improvement. This is a well-established fact for the Atari domain \cite{Schaul2016PrioritizedER}, but not quite explored in the areas of multi-task reinforcement learning or task-oriented language grounding.

\section{Conclusion}


In this work, we applied baseline methods of general deep reinforcement learning to the problem of task-oriented language grounding, where language input contains linear and non-linear order connectors.

We found that even baseline models can capture the semantics of linear and non-linear order connectors at the training instructions, but it is not enough to achieve high generalization performance even up to six sub-goals. The best results are achieved with the prioritized experience replay mechanism, which suggests its potential application to general multi-task reinforcement learning and task-oriented language grounding. And the most importantly, we found that training at both linear and non-linear order connectors helps to capture the semantics of the order connectors better when compared to just training on linear connectors -- the generalization performance increases in 2-3 times.

These findings suggest that we should look for a model that would generalize better even in such a simple setting: for instance, by introducing recurrence \cite{chaplot_gated-attention_2017, Mei2016ListenAA, hermann_grounded_2017} or hindsight experience replay \cite{andrychowicz_hindsight_2017}. But as the training performance is shown to be near perfect, it would be interesting to investigate the order connectors in visually rich environments or/and in the presence of other natural language instructions' semantics: multiple tasks, multiple referent attributes, and etc.

\bibliography{main}
\bibliographystyle{acl_natbib}
\end{document}